\documentclass[sigconf,natbib=false]{acmart}

\AtBeginDocument{%
  }

\setcopyright{acmlicensed}
\copyrightyear{2024}
\acmYear{2024}
\acmDOI{XXXXXXX.XXXXXXX}

\acmConference[]{3rd Workshop on End-to-End Customer
Journey Optimization at KDD 2024}{August 2024}{Barcelona, Spain}
\acmISBN{978-1-4503-XXXX-X/18/06}

\RequirePackage[
  datamodel=acmdatamodel,
  style=acmnumeric,
  ]{biblatex}

\begin{document}

\title[Dynamic Sequential Coupon Allocation Framework]{Optimizing Item-based Marketing Promotion Efficiency in C2C Marketplace with Dynamic Sequential Coupon Allocation Framework}

\author{Jie Yang}
\email{j-yang@mercari.com}
\affiliation{%
  \institution{Mercari, Inc.,}
  \city{Tokyo}
  \country{Japan}
}

\author{Padunna Valappil Krishnaraj Sekhar}
\email{pvkrajpv@mercari.com}
\affiliation{%
  \institution{Mercari, Inc.,}
  \city{Tokyo}
  \country{Japan}
}

\author{Sho Sekine}
\email{s-sekine@mercari.com}
\affiliation{%
  \institution{Mercari, Inc.,}
  \city{Tokyo}
  \country{Japan}
}

\author{Yilin Li}
\email{y-li@mercari.com}
\affiliation{%
  \institution{Mercari, Inc.,}
  \city{Tokyo}
  \country{Japan}
}


\renewcommand{\shortauthors}{Yang et al.}

\begin{abstract}
In e-commerce platforms, coupons play a crucial role in boosting transactions. In the customer-to-customer (C2C) marketplace, ensuring the satisfaction of both buyers and sellers is essential. While buyer-focused marketing strategies often receive more attention, addressing the needs of sellers is equally important. Additionally, the existing strategies tend to optimize each promotion independently, resulting in a lack of continuity between promotions and unnecessary costs in the pursuit of short-term impact within each promotion period.

We introduce a Dynamic Sequential Coupon Allocation Framework (DSCAF) to optimize item coupon allocation strategies across a series of promotions. DSCAF provides sequential recommendations for coupon configurations and timing to target items. In cases where initial suggestions do not lead to sales, it dynamically adjusts the strategy and offers subsequent solutions. It integrates two predictors for estimating the sale propensity in the current and subsequent rounds of coupon allocation, and a decision-making process to determine the coupon allocation solution. It runs iteratively until the item is sold. The goal of the framework is to maximize Return on Investment (ROI) while ensuring lift Sell-through Rate (STR) remains above a specified threshold. DSCAF aims to optimize sequential coupon efficiency with a long-term perspective rather than solely focusing on the lift achieved in each individual promotion. It has been applied for item coupon allocation in Mercari.


\end{abstract}




\maketitle

\begin{figure}
  \includegraphics[width=0.4\textwidth]{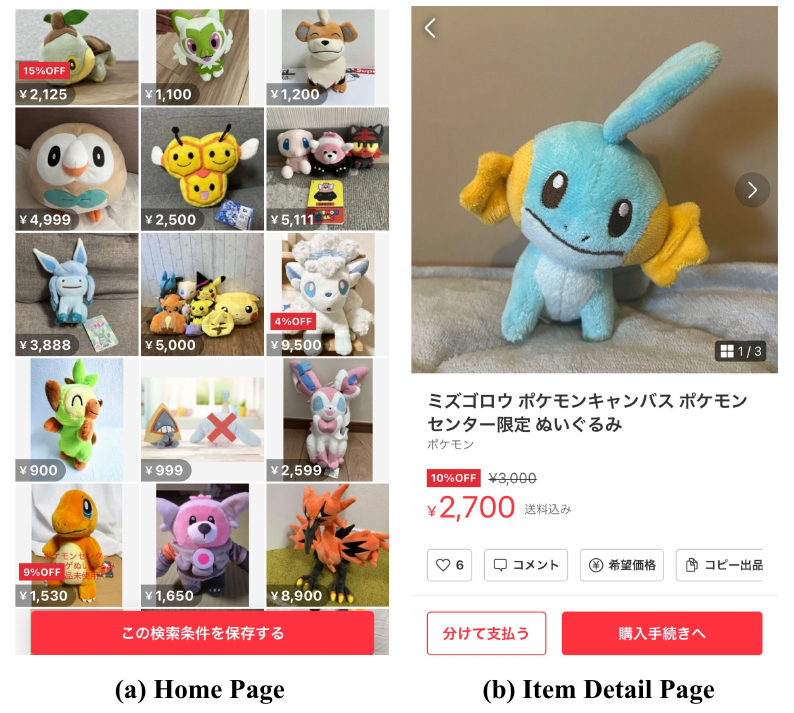}
  \caption{Item Coupon Display on Mercari. A discount badge is displayed when an item coupon is attached.}
  \Description{}
  \label{fig:experiments}
\end{figure}

\section{Introduction}
Coupon marketing is a business methodology used to attract, engage and reward customers by using different forms of incentives. In the C2C marketplace, engaging both buyers and sellers is essential for creating a robust buy-sell ecosystem. At Mercari, Japan’s largest C2C marketplace app, we distribute coupons to buyers as well as attach coupons to the items. To meet the satisfaction of both sides of buyers and sellers, the preceding work \cite{yang2022personalized} develops a personalized promotion decision-making framework on the buyer side, whereas this paper focuses on seller-side promotion optimization. Together, these approaches complement each other and aim to enhance the overall coupon marketing effectiveness and customer satisfaction of the two-sided marketplace.

In C2C marketplaces, millions of new items are listed daily. While new listings naturally attract buyer attention, they also dilute the visibility of existing listings, which have been listed for a while but remain unsold, reducing their chances of being sold. Different from product-based marketplaces, unsold listings can decrease sellers' satisfaction and potentially lead them to churn. Therefore, facilitating the sale of existing listings is crucial to retain sellers. When aiming to boost transactions through marketing strategies such as promotions and campaigns, most research focuses on the buyer side \cite{zhou2017bayesian, yamaguchi2021effect}. For sequential coupon allocation, research also mainly works on the buyer side \cite{li2020spending, xiao2019model}. Regarding the seller side, strategies typically involve adjusting exposure by reranking and updating item recommendation algorithms \cite{wu2009predicting}, whereas few focus on applying incentive campaigns, especially for the one-in-stock C2C case. However, incentive campaigns can prove effective in seller-focused initiatives, particularly when targeting specific seller segments or items. In our contexts, the emphasis lies in promoting existing listings that face reduced competitiveness relative to the new listings. Furthermore, incentive campaigns can optimize the entirety of the sales pipeline by integrating it into item ranking and recommendation systems.



To maintain sellers' satisfaction and encourage their continued use of the platform for subsequent listings, we need to boost the sales of their listings. Effective targeting strategy, includes three pivotal factors: target audience, coupon attachment timing and coupon configurations. In Mercari, when buyers express interest in items by clicking the like button, those items are saved to a designated like item list to ease future access. Nearly half of the purchases are made by customers who have liked items. Thereby, we position them as a crucial audience segment, with the action of liking serving as a pivotal precursor to the eventual transaction.

Regarding coupon attachment timing and coupon configurations, it is crucial to customize these aspects to correspond with the specific phases of the listing's life cycle stage. During the initial stage, when sellers create a new listing, there is typically a high organic purchase demand from customers. Therefore, it is advisable to attach coupons with small discount values or even refrain from attaching any coupons to minimize costs. However, as the listing progresses to a later phase and the initial surge in buyer interest subsides, more impactful coupons should be utilized to stimulate sales. If the initially attached coupon proves ineffective and fails to facilitate sales, a larger discount may be necessary in the later stage, leading to both increased costs and delays in transactions. Therefore, attaching the appropriate coupon at the right timing is crucial for achieving a significant business impact.

We introduce Dynamic Sequential Coupon Allocation Framework (DSCAF) to continuously enhance the effectiveness of item coupons throughout every stage of the item's lifecycle. DSCAF offers sequential suggestions on coupon configurations and timing to be applied to target items. In cases where the initial suggestion does not lead to sales, it adjusts the strategy and offers subsequent suggestions. The objective of this framework is to maximize ROI while maintaining lift STR above the business requirement. 

As shown in Figure \ref{fig:experiments}, DSCAF integrates two predictors for estimating sale propensity in the current and subsequent round of coupon allocation, along with a decision-making process that leverages both predictors. The following round predictor takes into account the item's status in the following round and utilizes the output of the model from the preceding round. We extract features from both the item and promotion viewpoints. Following the attachment of coupons to items, we dispatch notifications to the likers associated with each item to inform them about the promotional details. We iteratively execute the complete DSCAF pipeline until the item is sold.


In contrast to isolated marketing promotions that might lead to unnecessary costs in pursuit of short-term impact and lack continuity between promotions, DSCAF optimizes coupon efficiency through a long-term perspective by optimizing the sequence of marketing strategies.



 
\begin{figure}
  \includegraphics[width=0.45\textwidth]{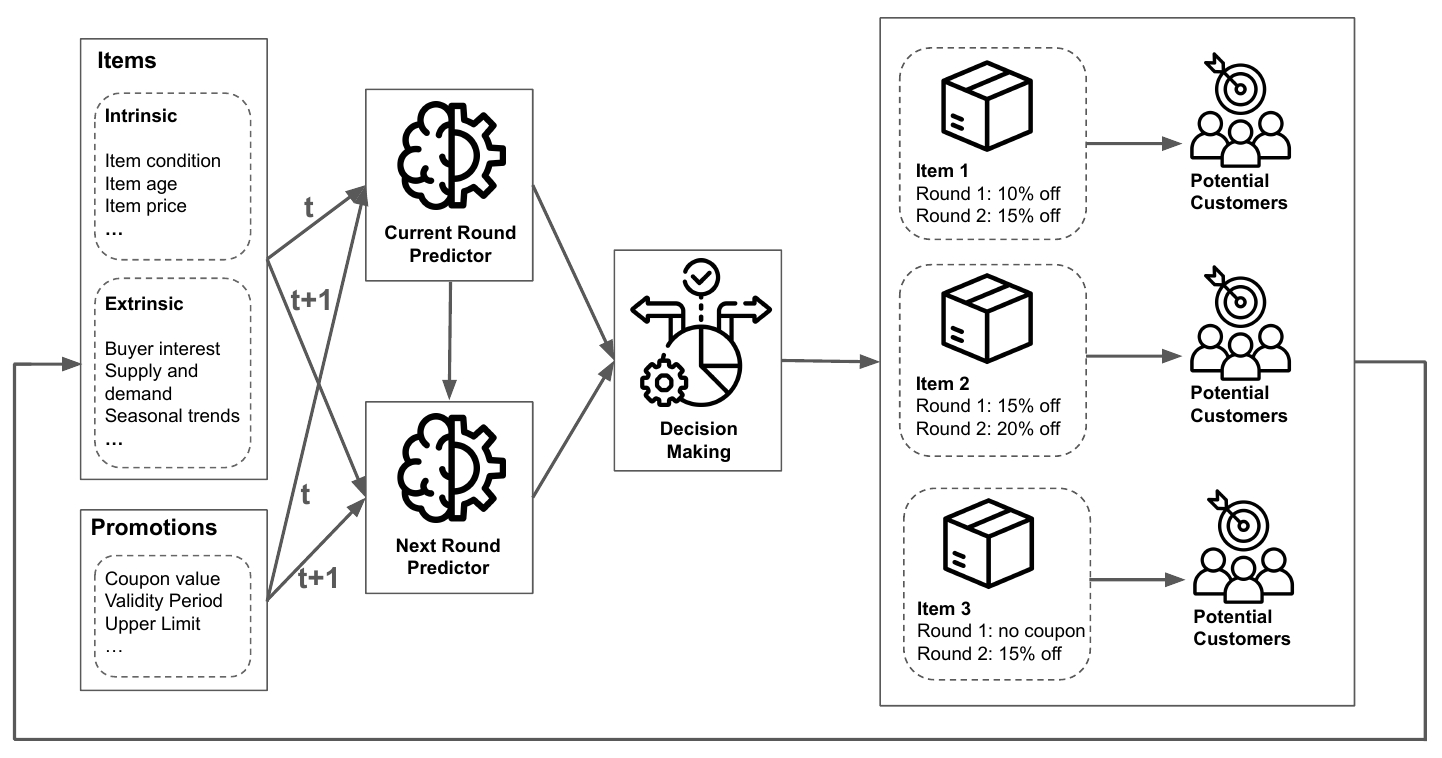}
  \caption{Dynamic Sequential Coupon Allocation Framework (DSCAF). }
  \Description{}
  \label{fig:experiments}
\end{figure}

\section{Methodology}
\label{Methodology}



DSCAF incorporates two predictors to estimate the sale propensity score for the current and next rounds. Each predictor estimates the Conditional Average Treatment Effect (CATE) on sales of items with different types of coupons attached during the corresponding promotion round. For both predictors, we employ S-learner \cite{kunzel2019metalearners} to estimate CATE.  


For training the first-round predictor, we utilize data obtained from a Randomized Controlled Trial (RCT) conducted during the initial coupon distribution round. Recognizing that not all items may be sold in the first round of promotions, we leverage the second predictor to estimate the sale propensity scores when coupons are attached in the subsequent round. This latter predictor is trained exclusively on the dataset encompassing items that remain unsold after the first round. During the second round, we solely have ground truth data for the items that remain unsold, lacking sales observations for items that are sold in the first round. However, during the inference phase, it is required to predict the sale propensity for every item in both rounds, including items that are akin to those that are sold during the first round. To mitigate bias stemming from non-random sampling in training the second-round predictor, we utilize the predictions of item remained propensity from the first-round predictor as Inverse Propensity Weights (IPW) \cite{seaman2013review} to weight the samples during the training of the second-round predictor.

In terms of item features, a combination of intrinsic and extrinsic factors is employed. Intrinsic factors include attributes inherent to the items, including its condition, age, price, etc., Extrinsic factors include external market conditions such as buyer interest, overall supply and demand dynamics and seasonal trends. Regarding coupon configurations, we consider the ensemble variations from discount amounts (e.g., 5\%, 10\%, 15\%), validity periods (e.g., 3 hours, 10 hours, 3 days) and coupon upper limit (such as 1000 JPY, 2000 JPY, 3000 JPY).

Once we have the predictions for both rounds, we compute the combined sale propensity score $p$ and ROI as follows:


\begin{equation}
\begin{aligned} 
  p^\dagger_{ij} & = f^\dagger(\boldsymbol{x^\dagger}_{ij}) \\
  p^\ddagger_{ik} & = f^\ddagger(\boldsymbol{x^\ddagger}_{ik}, \frac{1}{M}\sum_{j \in M } p^\dagger_{ij}) \\
  p_{ijk}  & = p^\dagger_{ij} + (1-p^\dagger_{ij})p^\ddagger_{ik} \\
  cost_{ijk}  & = \frac{p^\dagger_{ij}cost_{ij} + (1-p^\dagger_{ij})p^\ddagger_{ik}cost_{ik}}{p_{ijk}} \\
  ROI_{ijk} & = \frac{(p_{ijk} - p_{i}^*)LTV_{s(i)}}{cost_{ijk}}
\end{aligned} 
\end{equation}

Where $\boldsymbol{x^\dagger}, f^\dagger$ and $\boldsymbol{x^\ddagger}, f^\ddagger$ are the features and predictors of the first and second rounds respectively. $p^\dagger_{i;j}$ is the predicted sale propensity in the first round with choosing coupon $j$ for item $i$ whilst $p^\ddagger_{i;k}$ is the predicted sale propensity in the second round with choosing coupon $k$ and given the first round predictions among $M$ types of coupons. The latter part is used as weights for training. $cost_{ij}$ is the coupon cost of attaching coupon $j$ to item $i$. $p_{ijk}$ and $cost_{ijk}$ is the combined sale propensity and cost of choosing coupon $j$ and $k$ subsequently. $p_{i}^*$ is the sold propensity for item $i$ of not attaching coupons on either round. $LTV_{s(i)}$ is the LTV of the owner of item $i$. 

In the phase of making decisions about coupon allocation, we opt for the strategy with the highest $ROI_{ijk}$ for item $i$ among all feasible coupon allocation solutions where the lift sale propensity $p_{ijk}-p_{i}^*$ exceeds a predetermined threshold. This approach ensures that we prioritize achieving a high ROI while also maintaining a satisfactory gross sale value. Without constraints on lift sale propensity, there is a risk of choosing low-cost coupons to increase ROI, potentially resulting in lower lift STR and reduced gross sales value.



\section{Results and Applications}

\begin{figure}
  \includegraphics[width=0.47\textwidth]{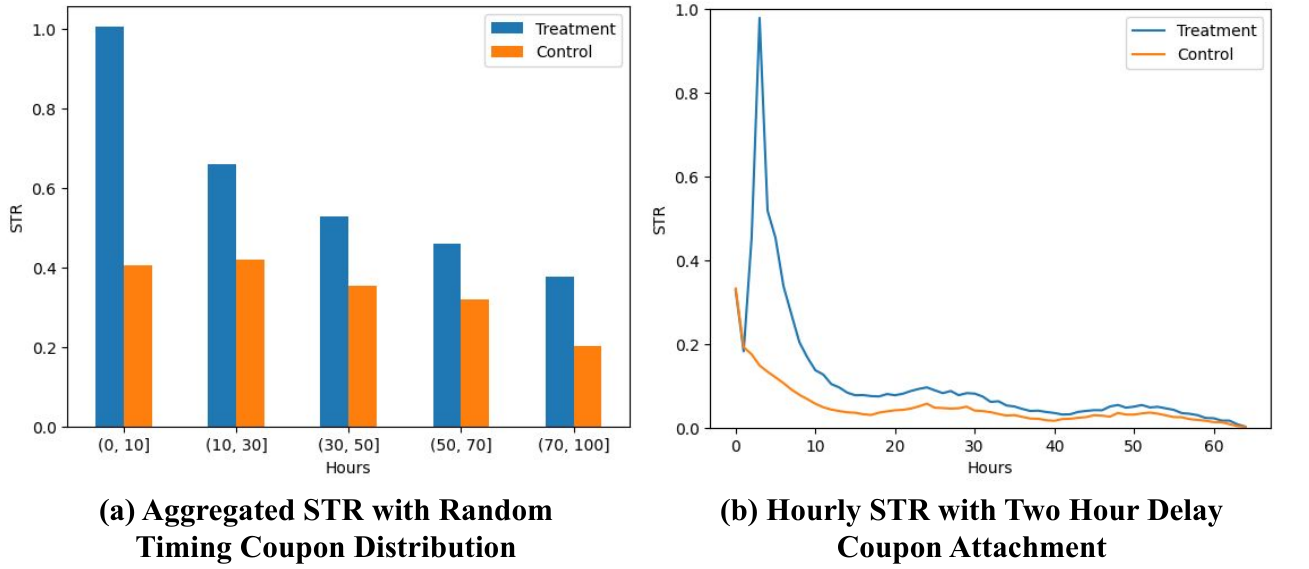}
  \caption{(a) STR with varying time delays following key actions. X-axis displays different hourly delay slots for coupon attachments; (b) Hourly STR of a two-hour delay distribution. X-axis is the transaction timing. Y-axis represents STR. }
  \Description{}
  \label{fig:time_delay}
\end{figure}

In this section, we compare three methods of item coupon distribution: (1) Select a distribution strategy randomly; (2) Optimize coupon distribution for each round of promotion independently; (3) Utilize DSCAF for coupon distribution.

To illustrate the significance of timing in coupon offerings, we analyze the lift STR by attaching coupons with different time delays after customers conduct key actions. Figure \ref{fig:time_delay}(a) demonstrates that items receiving an item coupon within ten hours after the key action have the highest lift STR. Beyond the ten-hour mark, buyer interest begins to wane, leading to a decrease in lift STR. This suggests that the timing is important to boost item purchases; the optimal time to target customers is several hours subsequent to their expression of interest in particular items.

In Figure \ref{fig:time_delay}(b), we observe a peak in lift STR immediately following the attachment of coupons. The lift remains significant for the initial ten hours post-coupon attachment, gradually decreasing thereafter. Beyond the twentieth hour, the difference between treatment and control groups becomes negligible, indicating that coupons lose their effectiveness beyond this point. It is important to note that attaching coupons immediately after customers conduct key actions showing intent will capture a larger audience. But this will also attach coupons to items with organic sales potential, increasing costs and reducing ROI.

Furthermore, we assess the Average Order Value (AOV) as it directly impacts the cost of coupons in scenarios where a percentage discount is offered. The trend of increasing AOV over time after the coupon is attached suggests that customers tend to purchase less expensive items shortly after receiving coupons and opt for more expensive items at a later stage. This insight prompts us to consider varying the validity period of coupons based on the price range of items. For cheaper items, a shorter validity period may be more suitable. If these items remain unsold, we can swiftly adapt our strategies and move on to the next promotion round. Conversely, for more expensive items, a longer validity period should be provided to allow customers ample time to make their purchasing decisions.

\begin{figure}
  \includegraphics[width=0.47\textwidth]{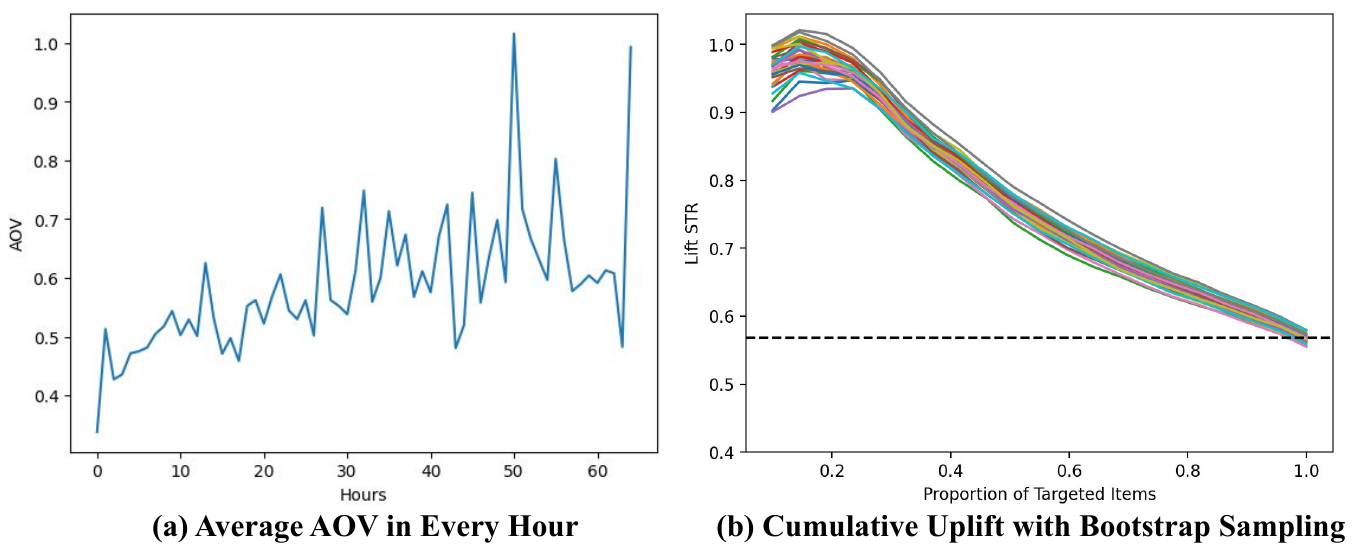}
  \caption{(a) AOV trends over hours after attaching item coupons; (b) Model performance evaluated by cumulative uplift with Bootstrap Sampling.}
  \Description{}
  \label{fig:aov}
\end{figure}

To train DSCAF models, we utilize a dataset containing 7 million items, with coupon types randomly assigned to the items as well as keeping a holdout group. We collect the data over a period of several months to mitigate seasonal biases. The model is fine-tuned using Optuna \cite{optuna_2019}, and the stability of predictions is assessed through Bootstrap Sampling. In the figure \ref{fig:aov} (b), we present the cumulative uplift \cite{gutierrez2017causal} on STR, with each curve representing a single sampling and the horizontal dotted line represents random sampling. The consistency of the decreasing trend in the curves and the lift from random sampling indicates that our model can identify item sale uplift propensity scores with stability. The absolute values in Figure \ref{fig:time_delay} and \ref{fig:aov} have been obscured by normalization. 


We conduct a comparison between method (2) and (3) based on ROI, training them on the same feature set with the same Optuna configuration. The results show that by combining subsequent rounds of promotions, we are able to increase ROI from 101\% to 173\% given the same lift STR constraint. This indicates that we can lower coupon costs and attain the same conversion levels by measuring conversions over a long-term period and optimizing the sequential item coupon allocation strategy accordingly. The framework has been rolled out for optimizing coupon distribution strategy for all existing items in Mercari.






\section{Conclusion}
In the C2C marketplace, engaging both buyers and sellers is essential for creating a robust buy-sell ecosystem. We introduce DSCAF to optimize item coupon allocation strategies as a solution for the seller side. It complements our preceding work \cite{yang2022personalized}, which focuses on optimizing promotion decision-making from buyer side. Furthermore, by applying the sequential coupon allocation strategy, we optimize coupon efficiency with a long-term perspective rather than focusing solely on individual promotions. This approach can reach the same transaction volume level with lower coupon costs. For future work, we plan to explore the integration of optimization strategies for both sellers and buyers.

\printbibliography

@String{Springer = "Springer-Verlag" }

@inproceedings{wu2009predicting,
  title={Predicting the conversion probability for items on C2C ecommerce sites},
  author={Wu, Xiaoyuan and Bolivar, Alvaro},
  booktitle={Proceedings of the 18th ACM conference on Information and knowledge management},
  pages={1377--1386},
  year={2009}
}

@article{zhou2017bayesian,
  title={Bayesian estimation of a dynamic model of two-sided markets: application to the US video game industry},
  author={Zhou, Yiyi},
  journal={Management Science},
  volume={63},
  number={11},
  pages={3874--3894},
  year={2017},
  publisher={INFORMS}
}

@article{yamaguchi2021effect,
  title={The effect of online C2C markets on new-product-purchasing behavior: an empirical analysis of Japanese selling apps},
  author={Yamaguchi, Shinichi},
  journal={SN Business \& Economics},
  volume={1},
  number={1},
  pages={26},
  year={2021},
  publisher={Springer}
}

@inproceedings{optuna_2019,
    title={Optuna: A Next-generation Hyperparameter Optimization Framework},
    author={Akiba, Takuya and Sano, Shotaro and Yanase, Toshihiko and Ohta, Takeru and Koyama, Masanori},
    booktitle={Proceedings of the 25th {ACM} {SIGKDD} International Conference on Knowledge Discovery and Data Mining},
    year={2019}
}

@article{yang2022personalized,
  title={Personalized Promotion Decision Making Based on Direct and Enduring Effect Predictions},
  author={Yang, Jie and Li, Yilin and Jobson, Deddy},
  journal={arXiv preprint arXiv:2207.14798},
  year={2022}
}

@article{seaman2013review,
  title={Review of inverse probability weighting for dealing with missing data},
  author={Seaman, Shaun R and White, Ian R},
  journal={Statistical methods in medical research},
  volume={22},
  number={3},
  pages={278--295},
  year={2013},
  publisher={Sage Publications Sage UK: London, England}
}

@inproceedings{gutierrez2017causal,
  title={Causal inference and uplift modelling: A review of the literature},
  author={Gutierrez, Pierre and G{\'e}rardy, Jean-Yves},
  booktitle={International conference on predictive applications and APIs},
  pages={1--13},
  year={2017},
  organization={PMLR}
}

@inproceedings{li2020spending,
  title={Spending money wisely: Online electronic coupon allocation based on real-time user intent detection},
  author={Li, Liangwei and Sun, Liucheng and Weng, Chenwei and Huo, Chengfu and Ren, Weijun},
  booktitle={Proceedings of the 29th ACM International Conference on Information \& Knowledge Management},
  pages={2597--2604},
  year={2020}
}

@inproceedings{xiao2019model,
  title={Model-based constrained MDP for budget allocation in sequential incentive marketing},
  author={Xiao, Shuai and Guo, Le and Jiang, Zaifan and Lv, Lei and Chen, Yuanbo and Zhu, Jun and Yang, Shuang},
  booktitle={Proceedings of the 28th ACM International Conference on Information and Knowledge Management},
  pages={971--980},
  year={2019}
}

@article{kunzel2019metalearners,
  title={Metalearners for estimating heterogeneous treatment effects using machine learning},
  author={K{\"u}nzel, S{\"o}ren R and Sekhon, Jasjeet S and Bickel, Peter J and Yu, Bin},
  journal={Proceedings of the national academy of sciences},
  volume={116},
  number={10},
  pages={4156--4165},
  year={2019},
  publisher={National Acad Sciences}
}

\appendix

\end{document}